\documentclass[conference,a4paper, 10pt]{APSIPA2025}
\usepackage{booktabs}
\usepackage[T5]{fontenc}              % T5 font encoding for Vietnamese
\usepackage{fontenc}[T1]{}

\usepackage{mathptmx}              % Times Roman text + math
\usepackage[utf8]{inputenc}           % Input encoding for Unicode source files
\usepackage[vietnamese,english]{babel} % Load Vietnamese language support

\usepackage{makecell}
\usepackage{amsmath}
\usepackage{tabularx} 
\usepackage{graphicx}
\usepackage{multirow}
\usepackage{threeparttable}
\usepackage[backend=biber,style=ieee,]{biblatex}
\usepackage{balance}
\addto\captionsenglish{\renewcommand{\figurename}{Fig.}}
\addto\captionsenglish{\renewcommand{\tablename}{TABLE}}

\usepackage{geometry}
\usepackage{fancyhdr}

\fancypagestyle{firststyle}{
  \fancyhf{}
  \fancyhead[C]{2025 Asia Pacific Signal and Information Processing Association Annual Summit and Conference (APSIPA ASC)}
}

\begin{document}

\title{VietLyrics: A Large-Scale Dataset and Models for Vietnamese Automatic Lyrics Transcription}

\author{
\authorblockN{
Nguyen Quoc Anh\authorrefmark{1} and
Bernard Cheng\authorrefmark{2} and
Kelvin Soh\authorrefmark{3}
}

\authorblockA{
\authorrefmark{1}
National University of Singapore, Singapore \\
E-mail: e1124714@u.nus.edu  Tel/Fax: +65-80706568}

\authorblockA{
\authorrefmark{2}
National University of Singapore, Singapore \\
E-mail: e1124449@u.nus.edu  Tel/Fax: +65-98557620}

\authorblockA{
\authorrefmark{3}
National University of Singapore, Singapore \\
E-mail: e1124460@u.nus.edu  Tel/Fax: +65-91186401}

}
\maketitle
\thispagestyle{firststyle}
\pagestyle{fancy}

\begin{abstract}
Automatic Lyrics Transcription (ALT) for Vietnamese music presents unique challenges due to its tonal complexity and dialectal variations, but remains largely unexplored due to the lack of a dedicated dataset. Therefore, we curated the first large-scale Vietnamese ALT dataset (VietLyrics), comprising 647 hours of songs with line-level aligned lyrics and metadata to address these issues. Our evaluation of current ASR-based approaches reveal significant limitations, including frequent transcription errors and hallucinations in non-vocal segments. To improve performance, we fine-tuned Whisper models on the VietLyrics dataset, achieving superior results compared to existing multilingual ALT systems, including LyricWhiz. We publicly release VietLyrics and our models, aiming to advance Vietnamese music computing research while demonstrating the potential of this approach for ALT in low-resource language and music.
\end{abstract}

\renewcommand{\thefootnote}{\fnsymbol{footnote}}
\footnotetext[1]{© 2025 IEEE. Personal use of this material is permitted. Permission from IEEE must be obtained for all other uses, in any current or future media, including reprinting/republishing this material for advertising or promotional purposes, creating new collective works, for resale or redistribution to servers or lists, or reuse of any copyrighted component of this work in other works.\\
This is the author’s accepted version of the paper: Nguyen Quoc Anh, Bernard Cheng, and Kelvin Soh, “VietLyrics: A Large-Scale Dataset and Models for Vietnamese Automatic Lyrics Transcription,” 2025 Asia Pacific Signal and Information Processing Association Annual Summit and Conference (APSIPA ASC), 2025. The final version is available on IEEE Xplore via DOI: https://doi.org/10.1109/APSIPAASC65261.2025.11249339}

\section{Introduction}

In the modern music industry, Automatic Lyrics Transcription (ALT) has become increasingly relevant and important. It describes the process of leveraging deep learning models to convert vocal recordings into written text, making song lyrics accessible and searchable. This enhances the user experience on streaming platforms, aids in music analysis, and supports subtitle creation for music videos. It also promotes inclusivity for the hearing impaired and non-native speakers. As the industry evolves, accurate lyrics transcription is essential for greater engagement and accessibility.

Despite advances in Western music computing research, Vietnamese music research lags behind, especially lyrics transcription, which remains mostly unexplored due to the lack of a large-scale, high-quality Vietnamese ALT dataset. Without technology like ALT or auto-tagging, Vietnamese media companies and streaming platforms lack the necessary tools to analyze, classify, and describe their content. This makes it difficult to build standardized datasets, hindering further research and innovation in the field, ultimately leaving Vietnamese music computing research stagnant.

This paper aims at tackling the task of automatic lyric translation for the Vietnamese language. Accurate transcription of Vietnamese is arguably more difficult than English due to additional language characteristics, such as changes in tonality leading to different meanings. Vietnamese also contains a multitude of regional dialects, each with unique pronunciation patterns and differences in vocabulary. Furthermore, changes in the speaker's pitch, accent, and emotionality can affect tonality, altering the diacritic mark, and thus the meaning of the transcribed word. For example, Tằm may sound similar to Tắm during singing, but they have completely different meanings. These complexities pose significant challenges in accurately transcribing Vietnamese vocals, especially when it is intertwined with melodies on the same track. 

In this work, these challenges are addressed by making the following key contributions:
\begin{itemize}
    \item The first large-scale Vietnamese ALT dataset, VietLyrics was constructed, which comprises 647 hours of songs with line-level aligned lyrics, supplemented with AI-predicted metadata such as gender and genre. The dataset is publicly released for research purposes, in compliance with Vietnamese copyright law.
    \item The Whisper \cite{radford2022robust} architecture was further fine-tuned using the VietLyrics dataset, significantly improving performance over existing multilingual ALT/ASR systems. Three model variants of different sizes are released to support further research and development in Vietnamese lyrics transcription.
\end{itemize}

\section{Related Works}
The following sections outline the current challenges in Vietnamese ALT and discuss the state-of-the-art solution (SOTA), detailed in Section \ref{subsection:lyrics_whiz}.
\begin{table}[h!]
  \centering
  \caption{Open-source datasets}
  \label{tab:opensource_datasets}
  \setlength{\tabcolsep}{4pt}
  \renewcommand{\arraystretch}{1.2}
  \begin{tabularx}{\linewidth}{X X c}
    \toprule
    \textbf{Dataset} & \textbf{Languages} & \textbf{Songs (Hours)} \\
    \midrule
    DALI-full \cite{DALI} & 30 (mostly EN) & 5,358 \\
    DALI-train \cite{DALI} & 1 (EN) & 3,913 \\
    MulJam \cite{zhuo2023lyricwhiz} & 6 (EN, FR, DE, ES, IT, RU) & 6,031 (381.9h) \\
    JamendoLyrics \cite{Durand_2023} & 4 (EN, ES, DE, FR) & 80 \\
    DSing30 \cite{Roa_Dabike-Barker_2019} & 1 (EN) & 4,324 (149.1h) \\
    MUSDB18 \cite{musdb18} & 1 (EN) & 150 (10h) \\
    \midrule
    VietLyrics-full (Ours) & 1 (VN) & 8,440 (647.1h) \\
    \bottomrule
  \end{tabularx}
\end{table}

\begin{table*}[htbp!]
\scriptsize

  \caption{Qualitative Analysis on Baseline LyricsWhiz's prediction}
  \label{tab:lyricswhiz_qualitative}
  \begin{tabular}{p{3cm} p{5cm} p{3cm} p{5cm}}
    \toprule
    \textbf{Ground Truth} & \textbf{Prediction} & \textbf{Source Song} & \textbf{Analysis}\\
    \midrule
     (No Vocals) & \textcolor{red}{Thành thành chuyện Thành thành; \newline Hãy subscribe cho kênh Ghiền Mì Gõ Để không bỏ lỡ những video hấp dẫn;...} & Ừ Thì Anh Sai - Thiên Tú & Long stretches of complete hallucination during no-vocals, due to the ASR model is not addapted into ALT\\
    \cmidrule[0.1pt](lr){1-4} % Light horizontal line
     (No Vocals) & \textcolor{red}{Lời Tắt Này Lời Tắt Này; \newline What the hell;... } & 365 Ngày Yêu (Remix) - Lương Gia Huy & Insertion of hallucinative random phrases\\
     \cmidrule[0.1pt](lr){1-4} % Light horizontal line
     \textcolor{green}{Nước} mắt em rơi cho cõi lòng em nắm tan & \textcolor{red}{Ngược} mắt em rơi cho cõi lòng em đã tan & Anh Biết Không Anh - Lưu Ánh Loan & Substitutions of single words that can reasonably be misheard\\
    \cmidrule[0.1pt](lr){1-4} % Light horizontal line   
    Chẳng lẽ ta mất nhau để em đây \textcolor{green}{khổ sầu}; \newline Thì ta đã \textcolor{green}{lìa} xa;...
     & Chẳng lẽ ta mất nhau để em đây \textcolor{red}{khô sâu}; \newline Thì ta đã \textcolor{red}{lịa} xa;...
 & Anh Biết Không Anh - Lưu Ánh Loan & Substitutions or deletion of a single word's diacritic that can reasonably be mis-transcribed due to regional accents.\\
     
    \bottomrule
  \end{tabular}
\end{table*}

\subsection{Dataset Scarcity} \label{section:lack_of_dataset}
% Table \ref{tab:opensource_datasets} details a non-exhaustive list of open source song-lyrics datasets.Given a lack of an established Vietnamese ALT dataset, our aim is to collect and curate our own custom Vietnamese song audio and lyric dataset. The collection process will be elaborated on in Section \ref{section:data_collection}.

Table \ref{tab:opensource_datasets} presents a non-exhaustive list of open-source song-lyric datasets. There is a lack of large-scale non-English song datasets in general, due to copyright law. These datasets circumvent this issue by collecting music exclusively under Creative Commons licenses or in the Public Domain \cite{DALI, Durand_2023, musdb18}, leveraging AI-generated lyrics \cite{zhuo2023lyricwhiz}, and incorporating karaoke vocals voluntarily shared by app users \cite{Roa_Dabike-Barker_2019}. None of these include Vietnamese, based on our findings, which leads to the lack of Vietnamese ALT models.

\subsection{Limitations in Existing Solutions}\label{subsection:lyrics_whiz}
% pho whisper
% whisper
% moises.ai

Although open source and commercial solutions are available for ALT or Automatic Speech Recognition (ASR) tasks, the general performance on Vietnamese music is error-prone. Various solutions were tested, including Whisper, LyricWhiz, PhoWhisper and one closed-source commercial API.  While all solutions resulted in some degree of substitution, insertion, deletion errors, along with long stretches of hallucination during periods with only melody and no vocals, ASR application Whisper stands out because it wrongly recognizes the audio as Thai or other languages regularly. 

% In fact, we can speculate that the closed-source API is Whisper-based, since its results also have the same signature Vietnamese hallucination \cite{whisper_issue} (possibly from the same data crawl from YouTube) similar to other Whisper-based applications.

% This project is largely inspired by the current state-of-the-art ALT framework, LyricWhiz\cite{zhuo2023lyricwhiz}. LyricWhiz combines leverages two models, 1. Whisper, an ASR model, and GPT-4, a large language model (LLM). Whisper functions as the "ear" by transcribing the audio to text, while GPT-4 acts as the "brain" by analyzing and selecting the most accurate lyrics. LyricWhiz has been shown to produce impressive accuracy across diverse languages and music genres, surpassing existing methods.

% However, there are multiple error instances when Vietnamese audio is used on the replicated baseline LyricsWhiz. The qualitative results of the baseline model are captured in Table \ref{tab:lyricswhiz_qualitative}. Most errors can be attributed to the inherent complexity of Vietnamese voices, or the fact that Whisper is natively an ASR model and not trained to adapt to ALT settings.

Among existing ALT frameworks, LyricWhiz \cite{zhuo2023lyricwhiz} is the current SOTA for multilingual ALT, leveraging both Whisper for transcription and GPT-4 for refining outputs. Whisper serves as the ``ear'' by converting audio into text, while GPT-4 functions as the ``brain'', selecting the most accurate lyrics. While LyricWhiz has demonstrated strong performance across various prominent languages, its effectiveness on Vietnamese lyrics is limited. The replicated baseline model struggles with Vietnamese audio, as shown in the qualitative results in Table \ref{tab:lyricswhiz_qualitative}. Note that other solutions also suffer from similar errors. Most errors stem from the inherent complexity of Vietnamese phonetics and the fact that Whisper, being primarily an ASR model, is not optimized for ALT tasks.

% One possible explanation for the observed discrepancies is that the limited amount of Vietnamese text in the original data set was used to train the Whisper model. This reinforces the point in Section \ref{section:lack_of_dataset} that a specialized dataset is required to be curated to achieve good ALT performance with under-represented languages like Vietnamese. 

% Explain the issues with qualitative results from baseline LyricsWhiz replication on Vietnamese songs

% \begin{table*}[h]
%   \caption{Performance Summary on Baseline LyricsWhiz}
%   \label{tab:performance_baseline_lyricswhiz}
%   \begin{tabular}{cl}
%     \toprule
%     Assessment & Examples \\
%     \midrule
%     Complete Hallucination & Output sentence: xx \textcolor{red}{xx} xx xx\\
%      & Ground Truth: xx \textcolor{green}{xx} xx xx \\
%     Single word  & Output sentence: xx \textcolor{red}{xx} xx xx\\
%     substitution & Ground Truth: xx \textcolor{green}{xx} xx xx \\
%     Missing/mistaken & Output sentence: xx \textcolor{red}{xx} xx xx\\
%     diacritics & Ground Truth: xx \textcolor{green}{xx} xx xx \\
%     \bottomrule
%   \end{tabular}
% \end{table*}

\section{Proposed Method}
\subsection{Dataset}
Around 647.1 hours of Vietnamese song audio with accompanying lyrics were scraped from \url{zingmp3.vn}, a Vietnamese music streaming site, to form the VietLyrics dataset. This dataset will be released publicly in the form of metadata, scraped links and scraping code, with author attributions to respect Vietnamese Intellectual Property Law. Quantitative and qualitative analysis shows that VietLyrics is diverse in genres, instruments, regional dialects, and gender. 
% Call it VietLyrics

\subsubsection{Data Collection \& Processing} \label{section:data_collection}
%  how was data collected
Out of 120,000 raw URLs collected by querying Vietnamese song titles using the \url{zingmp3.vn} search function, the dataset was filtered down to 17,000 entries. Heuristics were applied to exclude instrumental tracks, songs with English titles, titles containing numbers, and other irrelevant entries, ensuring the retained songs were predominantly in Vietnamese. For instance, searching for songs with numbers in their titles often resulted in tracks from other languages. To maximize dataset diversity, duplicate song names and potential cover versions were removed. A total of 12,758 songs and their metadata were successfully scraped without encountering significant web security challenges. Given that \url{zingmp3.vn} is region-restricted, a VPN should be used to access the platform from abroad. 

8,440 of 12,758 songs were found to have line-level lyric transcriptions that require further processing to ensure usability. This involved removing line breaks, punctuation, redundant timestamps, and introductory content from lyric transcriptions. Some songs were also identified as having incomplete transcriptions, in which the lyrics stopped before the final chorus repeats. To address this, the corresponding audio was trimmed to match the transcription length, plus an additional 10 seconds, ensuring no vocal segments were left without accompanying lyrics. Lastly, the sample rate for all audio tracks was standardized to 16 kHz.

Under Article 25, Clause 1(a) of the Vietnamese Intellectual Property Law (amended June 16, 2022), copyrighted musical works may be reproduced for academic and research purposes without permission or fees, provided the original creators are credited \cite{legal-disclaimer}. To respect copyright, we also scraped 2 other lyrics websites \cite{lyricvn.com, loibaihat.biz} for songwriters and artists for proper attributions, to the best of our abilities. We publicly release all metadata (including song names, links, and author attributions) and scraping code for academic research, allowing future studies to recreate the dataset, train, and benchmark for automatic transcription or other tasks.

\subsubsection{Data Analysis}
% The dataset covers a wide range of different song genres from more than 4,000 artists, as illustrated in Figure \ref{img:genre_dist}. However, metadata extraction from \url{zingmp3} results in a considerable amount of songs lacking genre information (classified as "unknown genre"). By analyzing the average words-per-minute (WPM) of the lyrics dataset, it shows that the average singing speed is more than twice as slow at an average WPM of 90.1 as compared to the average spoken words per minute (WPM) of 190\cite{lingopie_blog}. The underdeveloped field of Vietnamese music research lacks open-source tools like genre classification, or an agreed-upon ontology of Vietnamese music genres, so we infer VietLyrics on a set of top 50 common tags from MagnaTagATune \cite{magnatagatune}, using CLMR \cite{CLMR}. As the 95\% percentile of all tag prediction has only the confidence score of 0.22, CLMR did not find the Vietnamese music to be well-defined by any of the 50 MagnaTagATune tags. Therefore, Vietnamese music autotagging is an open problem for Vietnamese music AI research community.

\begin{figure}[htbp!]
  \centering
  \includegraphics[width=\linewidth, height=7cm]{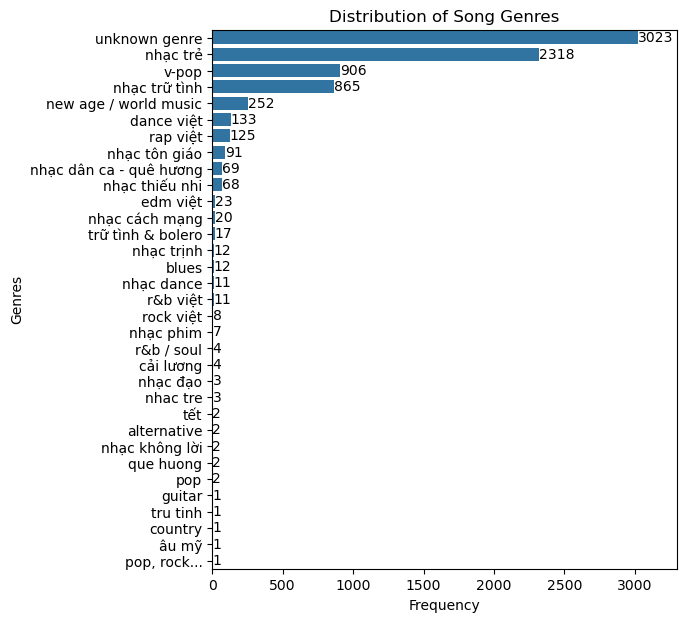}
    % \captionsetup{justification=centering}
  \caption{\centering Distribution of songs by genres. Extracted songs covers a wide range of genres, however a lot is missing.}
  \label{img:genre_dist}
\end{figure}

The dataset includes songs from over 4,000 unique artists, lasting on average 4.6 minutes/song. The average singing speed is found to be 90.1 Words Per Minute (WPM), less than half the Vietnamese spoken rate of 190 WPM \cite{lingopie_blog}. The dataset includes various genres, as shown in Fig. \ref{img:genre_dist}. However, metadata extraction from \url{zingmp3} alone resulted in many songs lacking genre information. Since Vietnamese music research lacks standardized genre classification tools or an agreed-upon genre ontology, Contrastive Learning of Musical Representations (CLMR) \cite{CLMR} was used to infer the top 50 MagnaTagATune tags \cite{magnatagatune}. The result was of weak alignment, with the 95th percentile confidence score reaching only 0.22. This suggests that this collection of Vietnamese music does not fit well within existing Western music's tag ontology, highlighting the need for further research in Vietnamese music genre classification.

Vietnamese dialects (which include both accents and variations in word choices, idioms, sentence structure, etc.) differ significantly across regions, a difference even more pronounced in singing, making accent classification valuable for music analysis. A pre-trained Wav2Vec2-base model \cite{Wav2Vec2, Wav2Vec2-base-vi} fine-tuned on Vietnamese Speech corpora \cite{2020-vlsp, LSVSC} was used to predict gender \cite{gender-classification} and dialect \cite{accent-classification}.

% pretrained on 13k hours of VLSP 2020 ASR audio \cite{2020-vlsp} and fine-tuned on the Novel Large-scale Vietnamese Speech Corpus \cite{LSVSC}, to classify gender \cite{gender-classification} and accent \cite{accent-classification}. 

\begin{figure}[h!]
  \centering
  \includegraphics[width=\linewidth, height=6.5cm]{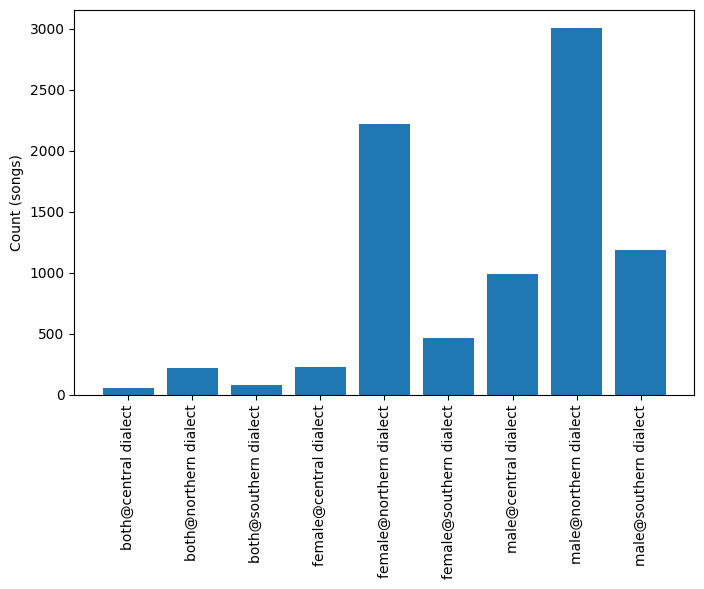}
  % \captionsetup{justification=centering}
  \caption{\centering Distribution of the dataset based on predicted genders and regional dialects.}
  \label{img:gender_dialect_dist}
\end{figure}

We extracted and ran inference on three 10-second vocal-only segments (isolated using Demucs v3 \cite{demux}) and determined the final classification through majority voting. For gender classification specifically, a ``both'' category was introduced for cases where at least one segment had a confidence score between 0.4 and 0.6, indicating ambiguity due to a unisex voice, absence of vocals, or a mix of male and female vocals. As shown in Fig. \ref{img:gender_dialect_dist}, the results indicate a distribution skewed towards Northern dialects, while Central dialects are underrepresented. Male-sung songs were more prevalent than those classified as ``female'' or ``both''. This data imbalance is hard to address, as gender and accent metadata were unavailable during data collection. (Note that this observed skew can also stem from Wav2Vec2 model bias.)

% Fig. \ref{img:song_duration_dist} shows that the song duration distribution is positively skewed around its mean of 4.6 minutes, with few outliers with long song durations. This average is relatively longer than the average duration of hit songs on mainstream music streaming platforms, such as Spotify with less than 3 minutes\cite{econlife}.

% \begin{figure}[h!]
%   \centering
%   \includegraphics[width=\linewidth, height=7cm]{images/song_duration_dist.png}
%   \caption{Song duration is normally distributed around mean of 4.6 minutes}
%   \label{img:song_duration_dist}
% \end{figure}

% The average words per minute (WPM) metric is also used as a proxy measure to understand the speed of the song in the curated dataset. Given that Vietnamese is one of the fastest spoken languages with an average words per minute (WPM) of 190\cite{lingopie_blog}, the average speed of the songs is more than twice as slow with average WPM at 90.1 as seen in Fig. \ref{img:wpm_dist}.

% The underdeveloped field of Vietnamese music research lacks open-source tools like genre classification, dialect classification, and accent recognition, limiting large-scale analysis of music datasets.

A qualitative examination revealed unique challenges, including lyrics in multiple languages (e.g., English, Mandarin in Latin characters), nonsensical vocalizations like ``là la la ...'' (\textit{Kết Thúc Vậy Sao} by Ngô Trác Lâm), and inconsistencies in whether background vocals are transcribed (\textit{Phút Giây Muộn Màng} by Bắc San Ho, \textit{Người Không Cần Yêu} by Khánh Đơn). These issues may require further processing to address.

% Using a Whisper-based model (which has an autoregressive language model as the transcriber) also poses challenges due to the uncommon language patterns in some songs, such as ancient Vietnamese literary styles and religious Buddhist texts, which are underrepresented in typical Internet scraped text datasets. Despite these difficulties, w

After using both manual qualitative sampling and quantitative exploration, the VietLyrics dataset is found to be sufficiently diverse, covering a large range of Vietnamese genres, traditional and modern instruments, regional music, and various dialects from across the country.

% \subsection{Source Separation}
% \label{source_separation}
% Hybrid Transformer Demucs (HT Demucs)\cite{rouard2022hybrid} is the SOTA music source separation model used to extract the vocal stem from the input audio files for the downstream ASR task. The hypothesis is that the removal of non-vocal stem would improve ALT performance as the musical elements in the audio would not be misinterpreted as speech. HT Demucs performs source separation by processing audio in both the temporal (waveform) and spectral (spectrogram) domains. It improves upon Hybrid Demucs\cite{defossez2021hybrid} by replacing the innermost layers with a cross-domain Transformer Encoder, allowing better integration of temporal and spectral information. 

% Fig. \ref{img:spectrogram_analysis} captures the effect of separation of the vocal stems on the input audio spectrogram using HT Demucs. The darker sections of the output spectrogram highlights the song sections of mainly musical accompaniments have been removed and replaced with relative silence.

% \begin{figure}[h!]
%   \centering
%   \includegraphics[width=\linewidth, height=7cm]{images/spectrogram_analysis.png}
%   \caption{Spectrogram comparison of Input Audio \& Source Separated Vocals}
%   \label{img:spectrogram_analysis}
% \end{figure}

\subsection{Modeling}
% Description/Diagram of model training pipeline
% Description of modifications (e.g. fine tuning, source separation, silence suppression, feature engineering, post-processing etc.)
% Model train-test split details
Since the current SOTA LyricWhiz model does not effectively handle Vietnamese songs, our approach is to fine-tune an ASR model end-to-end to serve as an ALT system. The Whisper model \cite{radford2022robust} was chosen as the starting point, as it is renowned for its SOTA performance in multilingual tasks and exceptional fine-tuning capabilities. \\
Three Whisper model variants of different sizes (small, medium and large-v2) were fine-tuned and made available as checkpoints on Hugging Face \cite{huggingface2024-whisper-large-v2}.

% 1. \textbf{openai/whisper-small}, 2. \textbf{openai/whisper-medium}, and 3. \textbf{openai/whisper-large-v2}. 

\subsubsection{Data Pre-processing}
Audio segmentation of songs is performed using the provided ground truth timestamps from the song lyrics to chunk both the audio and lyrics into 30-second segments. This is required as Whisper processes audio in 30-second chunks \cite{radford2022robust}.

To enhance training stability, audio and lyrics were sliced to keep only segments with corresponding ground truth transcription. This removes silences and parts without vocals, reducing dataset size and accelerating training while preserving meaningful content.

%, as illustrated in Figure~\ref{fig:audio-chunking} below.

% \begin{figure}[h]
%     \centering
%     \includegraphics[width=0.7\linewidth]{images/audio_chunking.png}
%     \caption{Ground Truth Lyrics}
%     \label{fig:audio-chunking}
% \end{figure}

% To further enhance audio quality and optimize the audio processing pipeline, we implemented an additional step of slicing the audio to retain only the segments containing lyrics, as indicated in the ground truth transcription. %Figure~\ref{fig:audio-chunking}
% In the original lyrics there are periods silence in different parts of the audio i.e. Chunk1 [00:04.00, 00:21.61, 00:23.99] and Chunk2 [00:30.84, 00:33.15]. This method effectively removes silences and irrelevant portions, significantly reducing the dataset size. The primary objective is to accelerate training while focusing on the meaningful segments of audio.  %Figure~\ref{fig:audio-chunk-remove-silence} illustrates how only the timestamps corresponding to lyrics are retained.

% \begin{figure}[h]
%     \centering
%     \includegraphics[width=0.7\linewidth]{images/audio_chunking_remove_silences.png}
%     \caption{Pre-Processed Ground Truth Lyrics}
%     \label{fig:audio-chunk-remove-silence}
% \end{figure}

\subsubsection{Fine-Tuning}
The dataset of 8,440 Vietnamese songs was divided into 7,440 training and 1,000 evaluation. All Whisper variants were fine-tuned using consistent hyper-parameters: batch\_size = 8, gradient\_accumulation\_steps = 2, learning\_rate = 1e-5, warmup\_steps = 500, and max\_steps = 5000. Training was conducted in FP16 precision to optimize computational efficiency. Our approach involved iterative fine-tuning, starting with smaller models (small and medium variants) to experiment with and refine our pre-processing techniques. After validating these improvements on the smaller models, we applied them to fine-tune the larger model, ensuring that the methodologies and pre-processing strategies generalized effectively across different model scales such as scaling from Whisper-small, to Whisper-medium and to Whisper-large-v2.

\section{Experiments and Results}

% Experiment Design details and its rationale
\subsection{Finetuned Models}

% Evaluation Metrics used (WER, CER etc.)
% We employed the LyricWhiz baseline as a first-pass zero-shot model, as the existing literature \cite{zhuo2023lyricwhiz} highlights its state-of-the-art performance in lyrics transcription at the time of writing.  However, the model's authors did not evaluate it on Vietnamese songs. We also benchmark PhoWhisper \cite{PhoWhisper}, an ASR model fine-tuned on 844 hours of Vietnamese audio, expecting it to perform better on Vietnamese audio. Consequently, when we tested the both zero-shot systems on our Vietnamese dataset, it produced a Word Error Rate (WER) of 49.22 for LyricWhiz and 38.99 for PhoWhisper-large, which falls short of acceptable standards to be used as a Automatic Lyrics Transcription model.

LyricWhiz was evaluated as a baseline for lyrics transcription, given its state-of-the-art performance in prior studies \cite{zhuo2023lyricwhiz}, though it had not been tested on Vietnamese songs. PhoWhisper \cite{PhoWhisper}, an ASR model fine-tuned on Vietnamese audio, was also assessed with the expectation of better performance on Vietnamese datasets. On VietLyrics test set, LyricWhiz produced a Word Error Rate (WER) of 49.22\%, while PhoWhisper-large achieved 38.3\%, shown in Table \ref{tab:model_performance}. However, both models fell short of the accuracy needed for effective Automatic Lyrics Transcription.

All Whisper models were fine-tuned on a dataset consisting of 7,440 training samples and 1,000 validation samples. The reported metrics were evaluated on the 1,000 validation samples. For context, Whisper was trained on 691 hours of Vietnamese ASR data, PhoWhisper was fine-tuned on 844 hours, and ours was fine-tuned on about 570 hours of songs. Our fine-tuned Whisper models were evaluated using two key metrics: Word Error Rate (WER) and Character Error Rate (CER). CER is particularly useful in accounting for cases where the model's predictions are close but incorrect due to minor diacritic errors, allowing such predictions to be considered as partially correct.

The results are summarized in Table~\ref{tab:model_performance}. As shown, scaling the model size significantly improves performance, with the best-performing Whisper-large-v2 model achieving a WER of 24.61\% and a CER of 17.14\%. The reported metrics demonstrate that Whisper-large-v2 is a robust ASR model suitable to be used for ALT for a low-resource language/music like Vietnamese. The relatively lower CER indicates that the model can accurately predict partial words, with errors arising from character or diacritic replacements.

\begin{figure}[h!]
  \centering
  \includegraphics[width=\linewidth]{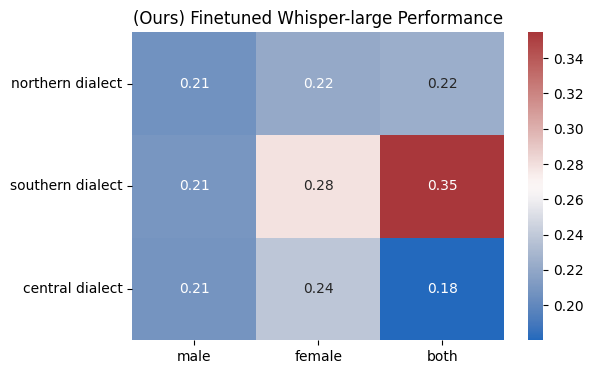}
% \captionsetup{justification=centering}
  \caption{ \centering Our Fine-tuned Whisper-large performance WER (\%), evaluated based on predicted genders and regional dialects.}
  \label{img:gender_dialect_perf_dist}
\end{figure}

Through qualitative analysis, Whisper-large-v2 was found to perform well on various instruments, music genres, and styles. As for gender and dialects, Whisper-large-v2 was found to have performed equally well in most cases, see Fig. \ref{img:gender_dialect_perf_dist}. Except for ``both'' gendered voices with Southern dialect, which we consider to be insufficient data to have any conclusive assertion, the model performs worse on female Southern dialect, a problem that needs to be addressed in future research.

We also experimented with training and evaluating only on vocals, separated using Demucs v3 \cite{demux}. However, the improvement observed after source separation was minimal for both models, with 1.3\% WER improvement for whisper-medium and 0.5\% WER improvement for whisper-large-v2. Given the negligible benefits and the added complexity of these extra pre-processing steps, the decision was made to exclude these steps from the final approach.

Lower-case evaluation was also found to produce better performance metrics compared to case-sensitive evaluation, with WER decreasing by up to 5\% on large, as shown in Table~\ref{tab:wer_case_comparison}. This intuitive result suggests that relying on the LLM-based transcriber for capitalization may not be the optimal approach. Instead, more effective methods for inferring line breaks could involve post-processing techniques, such as detecting silence between singing segments or leveraging the approaches introduced in Cífka et al. \cite{alt4humans}.

\begin{table}[h]
\centering
\caption{Model Performance \\ WER (\%) and CER (\%) Results.}
\label{tab:model_performance}
\setlength{\tabcolsep}{6pt}
\renewcommand{\arraystretch}{1.2}
\begin{tabularx}{0.9\linewidth}{|X|c|c|}
\hline
\textbf{Model} & \textbf{WER} & \textbf{CER} \\ \hline
Baseline: LyricWhiz & 49.22 & 38.55 \\ \hline
Baseline: PhoWhisper-large & 38.3 & 27.5 \\ \hline
Whisper-small (Fine-tuned) & 34.91 & 24.82 \\ \hline
Whisper-medium (Fine-tuned) & 26.42 & 17.03 \\ \hline
Whisper-large-v2 (Fine-tuned) & \textcolor{green}{24.61} & \textcolor{green}{17.14} \\ \hline
Whisper-medium (Demucs, Fine-tuned) & 25.12 & 16.34 \\ \hline
Whisper-large-v2 (Demucs, Fine-tuned) & \textcolor{green}{24.11} & \textcolor{green}{16.21} \\ \hline
\end{tabularx}
\end{table}

\begin{table}[htbp!]
\centering
\caption{Model Performance \\ WER (\%) Results for Case-sensitive vs Lowercase}
\label{tab:wer_case_comparison}
\setlength{\tabcolsep}{6pt}
\renewcommand{\arraystretch}{1.2}
\begin{tabularx}{0.9\linewidth}{|X|c|c|}
\hline
\textbf{Model} & \makecell{\textbf{WER}\\ \textbf{(Case-sensitive)}} & \makecell{\textbf{WER}\\ \textbf{(Lowercase)}} \\ \hline
Whisper-medium & 26.42 & \textcolor{green}{23.15} \\ \hline
Whisper-large-v2 & 24.61 & \textcolor{green}{20.52} \\ \hline
\end{tabularx}
\end{table}

\subsection{\texttt{<nospeech>} suppression}

% Since one of the major challenges in current ALT state-of-the-art models is hallucination induced by audio with no vocals, experiments were conducted to suppress \texttt{<nospeech>} tokens. Following the setup in \cite{radford2022robust}, tokens were filtered if the \texttt{<nospeech>} probability exceeded 0.6 and the log confidence probability was below -1. These experiments were evaluated on 100 randomly sampled songs from the 1000-song validation set.

% PhoWhisper was tested against various configurations. Only Vocals tracks were extracted using Demucs. As shown in Table~\ref{tab: silence_supression}, on PhoWhisper-small, a combination of source-separated audio and \texttt{<nospeech>} suppression yielded the best, albeit modest, results. For Whisper-large size, testing on raw data without suppression yields the best performance, with our fine-tuned Whisper-large-v2 significantly outperforming others. 

Since hallucination from non-vocal audio is a major challenge in ASR models, experiments were conducted to suppress \texttt{<nospeech>} tokens. Following the original Whisper paper \cite{radford2022robust}, tokens were filtered when the \texttt{<nospeech>} probability exceeded 0.6 and the log confidence probability was below -1. These experiments were evaluated on 100 randomly sampled songs from the 1,000-song validation set. PhoWhisper was tested under various configurations, with vocals-only tracks extracted using Demucs. As shown in Table~\ref{tab: silence_supression}, for PhoWhisper-small, combining source-separated audio with \texttt{<nospeech>} suppression provided the best performance. However, for Whisper large, testing on raw audio without suppression performed best, with our fine-tuned Whisper significantly outperforming others.

This outcome is consistent with the findings in Cífka et al. \cite{alt4humans}, where the transcriptions from only vocals are less accurate than those from raw data for large models. This is preferable, as it enables deploying an end-to-end model without dependence on source separation performance. Furthermore, qualitative reviews of our finetuned Whisper's inferences found no instances of long hallucinations, resulting in the discontinuation of further silence suppression experiments.

\begin{table}[h!]
\centering
\caption{Silence Suppression Performance \\ WER (\%) and CER (\%) results for Lowercase.\\ \textbf{NS} means <nospeech> token is suppressed.}
\label{tab: silence_supression}
\setlength{\tabcolsep}{6pt}
\renewcommand{\arraystretch}{1.2}
\begin{tabularx}{0.95\linewidth}{|X|X|c|c|c|}
\hline
\textbf{Model} & \textbf{Input} & \textbf{NS} & \textbf{WER} & \textbf{CER} \\ \hline
\makecell[l]{Whisper-large-v2 \\ (Ours)} & Raw Data & No & \textcolor{green}{20.1} & \textcolor{green}{16.0} \\ \hline
% \makecell[l]{Wp-large-v2\\(Ours)+EC} & Raw Data & No & 26.7 & 22.2 \\ \hline
PhoWhisper-large & Raw Data & No & 38.3 & 27.5 \\ \hline
PhoWhisper-large & Only Vocals & Yes & 41.8 & 31.5 \\ \hline
PhoWhisper-small & Raw Data & No & 54.2 & 38.4 \\ \hline
PhoWhisper-small & Raw Data & Yes & 54.4 & 39.1 \\ \hline
PhoWhisper-small & Only Vocals & No & 50.2 & 35.3 \\ \hline
PhoWhisper-small & Only Vocals & Yes & \textcolor{blue}{48.7} & \textcolor{blue}{34.5} \\ \hline
\end{tabularx}
\end{table}

\section{Conclusion}

We collected VietLyrics, a large-scale diverse dataset of Vietnamese music with lyrics, and fine-tuned the Whisper models \cite{radford2022robust} to develop an Automatic Lyrics Transcription system. Our best performing model, Whisper-large-v2, achieved a WER of 24.61\% for case-sensitive lyrics and 20.52\% for lowercase lyrics, demonstrating its effectiveness in transcribing Vietnamese songs of various dialects, genres and genders. We publicly release the VietLyrics dataset \cite{VietLyricsDataset} and models \cite{whisper-large-model, whisper-medium-model, whisper-small-model}, hoping that this work inspires further research and improvements in Automatic Lyrics Transcription and music computing in general, for the Vietnamese language and music.

% We have made the Whisper-small, Whisper-medium and Whisper-large-v2 models publicly available on the HuggingFace 
% \href{https://huggingface.co/xyzDivergence/whisper-large-v2-vietnamese-lyrics-transcription}{HuggingFace Hub} \cite{whisper-small-model}, \cite{whisper-large-model}, \cite{whisper-medium-model} 
% and plan to share the dataset through songs' links. 

\section{Acknowledgment}
We would like to thank Zhao Junchuan and Nguyen Thi Minh Thu for their invaluable help and support throughout this research.

% \begin{thebibliography}{1}

% \bibitem{1}
% G.~Eason, B.~Noble, and I.~N.~Sneddon, ``On certain integrals of
% Lipschitz-Hankel type involving products of Bessel functions,''
% \emph{Phil. Trans. Roy. Soc. London,} vol. A247, pp. 529-551, April
% 1955.

% \bibitem{2}
% J.~Clerk~Maxwell, \emph{A Treatise on Electricity and Magnetism,}
% 3$^{\rm rd}$ ed., vol. 2. Oxford: Clarendon, 1892, pp.68-73.

% \bibitem{3}
% I.~S.~Jacobs and C.~P.~Bean, ``Fine particles, thin films and exchange
% anisotropy,'' in \emph{Magnetism,} vol. III, G.T. Rado and H. Suhl,
% Eds. New York: Academic, 1963, pp. 271-350.

% \bibitem{4}
% K.~Elissa, ``Title of paper if known,'' unpublished.

% \bibitem{5}
% R.~Nicole, ``Title of paper with only first word capitalized,''
% \emph{J. Name Stand. Abbrev.,} in press.

% \bibitem{6}
% Y.~Yorozu, M.~Hirano, K.~Oka, and Y.~Tagawa, ``Electron spectroscopy
% studies on magneto-optical media and plastic substrate interface,''
% \emph{APSIPA Transl. J. Magn. Japan,} vol. 2, pp. 740-741, August 1987
% [\emph{Digests 9$^{\rm th}$ Annual Conf. Magnetics Japan,} p. 301,
% 1982].

% \bibitem{7}
% M.~Young, \emph{The Technical Writer's Handbook.} Mill Valley, CA:
% University Science, 1989.

% \end{thebibliography}

\printbibliography
\balance

\end{document}